\documentclass{article}
\usepackage{spconf,amsmath,graphicx}
\usepackage{url}
\usepackage[breaklinks,colorlinks]{hyperref}
\usepackage{multirow}
\usepackage{array}
\usepackage{cleveref}
\usepackage{amssymb}
\usepackage{makecell}
\usepackage{color} 

\Crefname{figure}{Fig.}{Figs.}

\title{BioVL-QR: Egocentric Biochemical Vision-and-Language Dataset \\ Using Micro QR Codes}
\name{\begin{tabular}{c}Tomohiro Nishimoto$^1$, Taichi Nishimura$^1$, Koki Yamamoto$^1$, Keisuke Shirai$^1$, Hirotaka Kameko$^1$,\\  Yuto Haneji$^1$, Tomoya Yoshida$^1$, Keiya Kajimura$^1$, Taiyu Cui$^1$, Chihiro Nishiwaki$^2$,\\ Eriko Daikoku$^2$, Natsuko Okuda$^2$, Fumihito Ono$^2$, Shinsuke Mori$^1$\end{tabular}}
\address{$^1$Kyoto University, $^2$Osaka Medical and Pharmaceutical University}
\begin{document}
%
\maketitle
\begin{abstract}
This paper introduces BioVL-QR, a biochemical vision-and-language dataset comprising 23 egocentric experiment videos, corresponding protocols, and vision-and-language alignments.
A major challenge in understanding biochemical videos is detecting equipment, reagents, and containers because of the cluttered environment and indistinguishable objects.
Previous studies assumed manual object annotation, which is costly and time-consuming. 
To address the issue, we focus on Micro QR Codes.
However, detecting objects using only Micro QR Codes is still difficult due to blur and occlusion caused by object manipulation.
To overcome this, we propose an object labeling method combining a Micro QR Code detector with an off-the-shelf hand object detector.
As an application of the method and BioVL-QR, we tackled the task of localizing the procedural steps in an instructional video.
The experimental results show that using Micro QR Codes and our method improves biochemical video understanding.
Data and code are available through \url{https://nishi10mo.github.io/BioVL-QR/}.

\end{abstract}
\vspace{-0.5em}
\section{Introduction}
\vspace{-0.5em}
\label{sec:intro}
Low reproducibility is critical in science.
As Baker reported~\cite{baker2016nature}, in biochemistry, more than 80\% of scientists have failed to reproduce other scientists' experiments, and more than 60\% have failed to reproduce even their own experiments.
Improving reproducibility is important to make scientific findings universally applicable.

Scientific experiments contain multimodality in nature.
In real-world scenarios, a typical approach to reproducing experiments allows researchers to read protocols and execute the experiments. 
However, this process is prone to human errors.
Therefore, constructing multimodal systems that support experimental execution can be a promising way to enhance reproducibility.
For example, it is useful to have a system that detects human errors and verbalizes how to recover them to prevent researchers from failing to reproduce experiments.
To construct systems of this kind, vision-and-language technologies, such as step localization~\cite{naim-2015, finebio, stepformer}, are effective approaches.
Step localization allows researchers to visually verify the correct procedure.

\begin{figure}[t]
    \centering
    \includegraphics[width=1\linewidth]{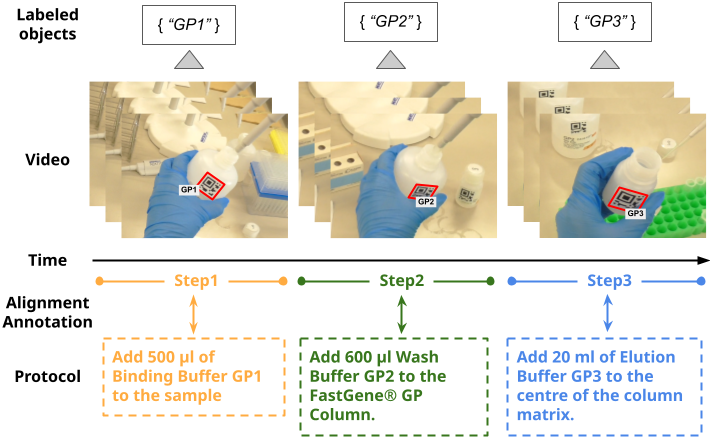}
    \caption{Overview of BioVL-QR containing experiment videos and corresponding protocols. We label objects appearing in the videos using Micro QR Codes.}
    \label{fig:abst_biovlqr}
\end{figure}

Previously, a few researchers attempted to focus on the vision-and-language tasks in the field of biochemistry~\cite{finebio, biovl2, probio}.
The key challenge in such tasks is detecting equipment, reagents, and containers because the lab environment is scattered by filling objects on the table, and some objects are even indistinguishable.
For example, it is difficult to distinguish tubes that look identical but contain different reagents.
In \Cref{fig:abst_biovlqr}, distinguishing between ``GP1'' and ``GP2'' is challenging due to their identical appearance. However, correctly identifying them is crucial for real-world applications, as their chemical properties differ significantly.
Previous studies~\cite{naim-2015,biovl2} assumed that such visually-indistinguishable objects were manually annotated and given for downstream tasks, but this is costly.

To tackle this, we focus on Micro QR Codes~\cite{microqr} to detect objects automatically.
In this paper, we introduce BioVL-QR, a biochemical vision-and-language dataset containing 23 egocentric experiment videos, protocols, and alignment annotation, where Micro QR Codes are attached (\Cref{fig:abst_biovlqr}). In BioVL-QR, the objects appearing in the videos are attached with Micro QR Codes.
In our preliminary studies, we found that detecting objects only using Micro QR Codes is still difficult because the researchers manipulate objects, frequently causing blur and occlusion.
To address this, we also propose a novel object labeling method by combining a Micro QR Code detector and an off-the-shelf hand object detector~\cite{hobj}.
Based on BioVL-QR, we conducted a task of step localization, localizing the steps described in procedural texts within an instructional video.
The experimental results show that our approach improves the understanding of biochemical videos.

In summary, our main contributions are: (1) the BioVL-QR dataset, (2) an object labeling method, and (3) baseline methods for step localization that use Micro QR codes and the proposed object labeling method.

\section{Related work}
\vspace{-0.5em}
\label{sec:relwork}
Egocentric vision-and-language datasets are important for advancements in fields such as AR~\cite{HoloAssist} and robotics~\cite{bhateja2024robotic}. Large-scale egocentric vision-and-language datasets are actively being created~\cite{ego4d, assembly101}.
These large-scale datasets do not include biochemistry videos primarily due to the difficulty of domain-specific annotation.
Specifically, annotating the experimental procedures and labeling the objects in the videos require a high level of expertise in the relevant scientific domain. 
This makes the annotation process both time-consuming and resource-intensive.

In the field of biochemistry, several small-scale egocentric vision-and-language datasets were proposed~\cite{biovl2, finebio}.
Similar to BioVL-QR, BioVL2~\cite{biovl2} and FineBio~\cite{finebio} include egocentric videos, corresponding protocols, and annotations of objects in the videos. However, in these datasets, the names of objects in the videos are manually annotated, which requires significant effort.
To address this issue, we propose an automated annotation method using Micro QR Codes and object labeling.

\section{BioVL-QR Dataset}
\vspace{-0.5em}
\label{sec:data}
This section introduces BioVL-QR.
First, we describe video recording (\Cref{ssec:video_rec}).
Then, we explain the alignment annotation (\Cref{ssec:align_annot}). Finally, we report the dataset statistics (\Cref{ssec:stat}).

\subsection{Video Recording}
\vspace{-0.5em}
\label{ssec:video_rec}
\textbf{Setting.}
We asked a student who majors in biochemistry to wear a head-mounted camera (Panasonic HX-A50) to record experiments. We recorded videos at 30 fps and 4K (3,840$\times$2,160) resolution to improve the detection performance of Micro QR Codes.
Note that we attached Micro QR Codes to as many objects relevant to the experiment, including equipment, reagents, and containers as possible.
Our preliminary studies show that Micro QR Codes smaller than 1cm$^2$ are difficult to recognize. Therefore, we prepared for them larger than 1cm$^2$.
In this study, we used three sizes of QR Codes: 1cm, 2cm, and 3cm according to the size of the objects.
When recording the videos, we asked the researcher to conduct the experiments as they normally would. Additionally, we requested that they avoid unusual operations, such as intentionally holding objects in a way that makes the Micro QR Codes clearly visible in the video.
For our dataset, we chose four basic experiments frequently performed in biochemistry: DNA extraction, making an agarose gel, electrophoresis, and DNA purification. 
Six videos were recorded for experiments other than DNA purification, but only five were recorded for DNA purification due to a recording failure.

\noindent\textbf{Data preprocessing.}
We removed the audio from the videos for privacy reasons. Additionally, we blurred faces that appeared in the videos when they were facing the camera. Furthermore, we manually removed parts of videos where the researcher did not operate for more than 30 seconds, such as waiting for a centrifuge to finish.
After data processing, the total duration of the experiment videos is 2.09 hours.

\begin{table}[t]
\centering
\caption{Annotation agreement rate.}
\label{tabl:tIoU}
    \scalebox{0.9}{
    \begin{tabular}{l|r}
    \hline
    Task               & \multicolumn{1}{c}{tIoU{[}\%{]}} \\
    \hline
    DNA extraction     & 74.3                             \\
    electrophoresis    & 83.6                             \\
    making agarose gel & 82.3                             \\
    DNA purification      & 71.2    \\
    \hline
    \end{tabular}
    }
\end{table}

\begin{figure*}[t]
    \centering
    \includegraphics[width=0.9\textwidth]{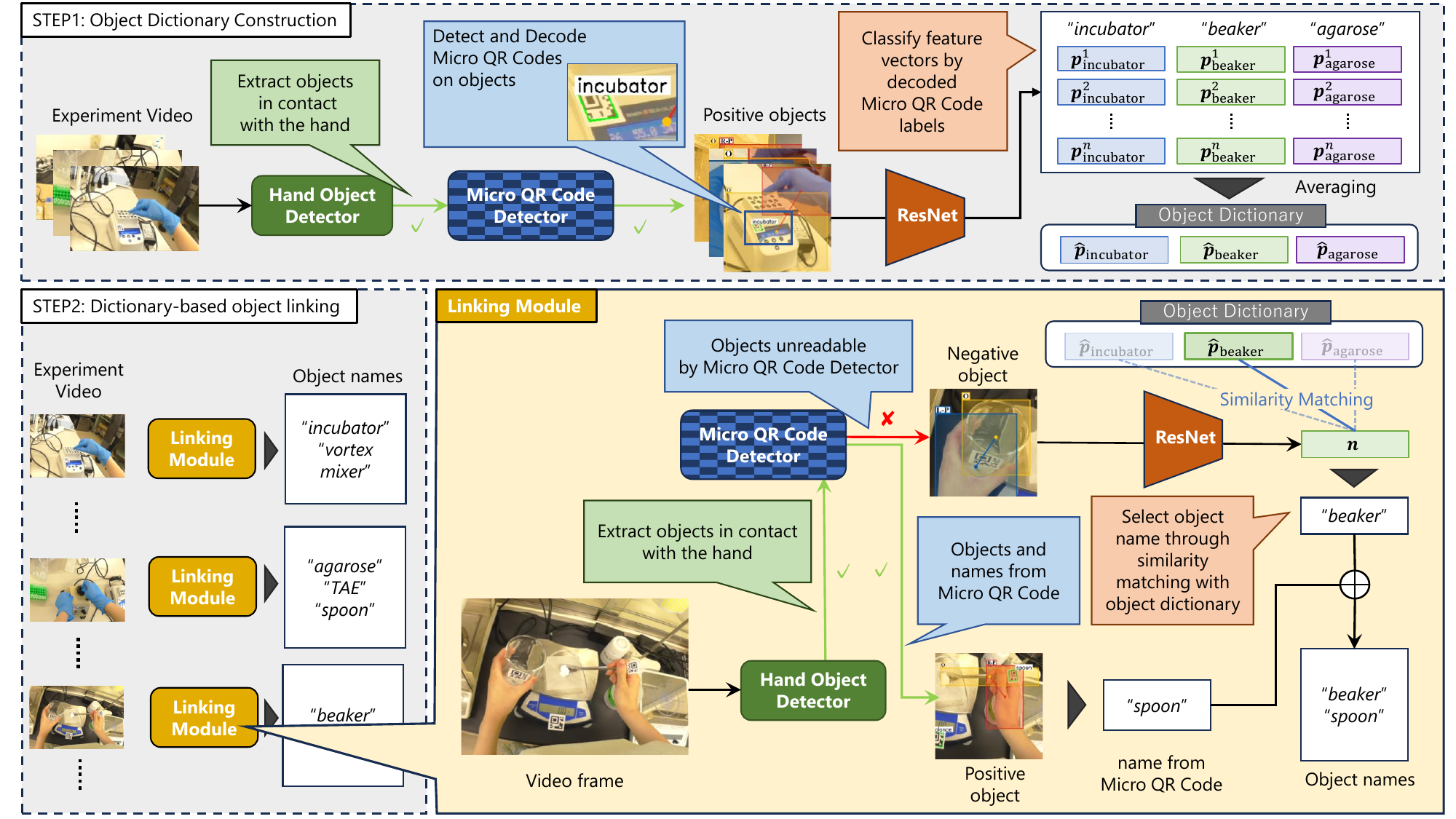}
    \caption{Overview of our object labeling process, which consists of two steps: object dictionary construction (step 1) and dictionary-based object linking (step 2). Step 1 constructs an object dictionary by associating object names obtained from Micro QR Codes with their visual feature vectors. Step 2 links hand-interacted objects in the video to their names using an object dictionary.}
    \label{fig:auto_ann}
\end{figure*}

\subsection{Alignment Annotation}
\vspace{-0.5em}
\label{ssec:align_annot}
Following the existing vision-and-language datasets~\cite{youmakeup_2019, crosstask_2019}, we provide alignment annotation, which aligns event spans in videos with steps in procedural texts.
We annotated this alignment as the following two processes.
First, we split sentences based on actions. For example, ``Discard the flow-through and place the FastGene GP Column back into the Collection tube'' was split into ``Discard the flow-through'' and ``Place the FastGene GP Column back into the Collection tube''.
Then, annotators were asked to provide start and end timestamps and link them with the split sentences.

\noindent\textbf{Annotation agreement.}
To ensure the annotation quality, we assessed the annotation agreement by hiring a different annotator.
Specifically, we compared these two annotation results and computed temporal Intersection over Union (tIoU), a metric measuring the temporal overlap between two events~\cite{tvretrieval2020}.
\Cref{tabl:tIoU} shows that tIoU exceeds 70\% in all four types of experiments, indicating high annotation quality~\cite{tvretrieval2020, gao2021fast}.

\subsection{Statistics}
\vspace{-0.5em}
\label{ssec:stat}
\begin{table}[t]
\centering
\caption{Statistics of the dataset in each modality.}
\label{tab:video_language_stats}
\scalebox{0.75}{
    \begin{tabular}{c|rr|rr}
    \hline
    \multirow{3}{*}{Task}               & \multicolumn{2}{c|}{language side}                                 & \multicolumn{2}{c}{video side}                               \\ \cline{2-5} 
                       & \multicolumn{1}{c}{\makecell{Avg. steps \\per protocol}} & \multicolumn{1}{c|}{\makecell{\#words \\per step}} & \multicolumn{1}{c}{\makecell{Avg. duration \\(sec)}} & \multicolumn{1}{c}{\#objects} \\ \hline
    DNA extraction     & 5.0                         & 7.6                                  & 143.2                        & 7.0                           \\
    electrophoresis    & 7.0                         & 9.9                                  & 409.1                        & 10.0                          \\
    making agarose gel & 6.3                         & 7.2                                  & 395.6                        & 5.6                           \\
    DNA purification   & 19.0                        & 9.5                                  & 369.2                        & 11.0                          \\ \hline
    \end{tabular}
    }
\end{table}
We provide statistics of BioVL-QR from two perspectives: the language side and the video side.

\noindent\textbf{Language side.}
\Cref{tab:video_language_stats} (left) shows the language side statistics, indicating the diversity of the protocols. DNA purification has the longest protocol, while DNA extraction has the shortest one. DNA purification includes many steps with short execution times, such as ``Place a FastGene® GP Column into a Collection Tube.'' As a result, it has significantly more steps than other experiments.

\noindent\textbf{Video side.}
\Cref{tab:video_language_stats} (right) shows the video side statistics, demonstrating that the videos are also diverse because the durations and objects are different among experiments. Per unit time, DNA extraction involves the highest number of objects, while making agarose gel involves the fewest.

\section{Object labeling based on QR Codes}
\vspace{-0.5em}
\label{sec:labeling}
In this section, we describe the object labeling method (\Cref{ssec:labeling_method}) and its evaluation (Section \ref{ssec:labeling_eval}).

\subsection{Method}
\vspace{-0.5em}
\label{ssec:labeling_method}
\noindent\textbf{Labeling process overview.}
We automatically label the objects appearing in experiment videos based on Micro QR Codes.
Preliminarily, we attempted to label objects using only Micro QR Codes, but we could not get satisfactory accuracy due to blur and occlusion caused by object manipulation.
On the other hand, the hand object detector~\cite{hobj} performs well in detecting hands and interacted objects, yet it cannot obtain the object names because it only detects objects interacting with hands.

Based on this observation, we propose a novel object labeling method that combines Micro QR Codes with a hand object detector (\Cref{fig:auto_ann}).
It consists of two processes: (1) object dictionary construction and (2) dictionary-based object linking.
The key idea is to connect the hand-interacted objects with object names via the pre-constructed dictionary, which consists of pairs of object names and appearance vectors.

\noindent\textbf{Object dictionary construction.}
In this process, we construct the object dictionary, which consists of object names as keys and their corresponding visual feature vectors as values.
For each video, we sample frames at 10 fps and apply the hand object detector~\cite{hobj} to extract the bounding boxes of objects in contact with hands.
These are forwarded to the Micro QR Code detector to read object names.
After this process, some Micro QR codes are successfully read, while others are not.
The former successful ones are called ``positive objects'' in this paper, and the latter are called ``negative objects.''
Then, we extract the appearance feature vectors $\mathbf{p}_{name}^{i}$ by encoding the image regions corresponding to the bounding boxes using ResNet-50~\cite{ResNet}, where $name$ represents the object names obtained from the Micro QR Code detector and ${i}$ represents the index of vectors.
We aggregate the object-wise feature vectors $\hat{\mathbf{p}}_{name}$ by averaging them and obtaining the object dictionary.

\noindent\textbf{Dictionary-based object linking.}
Based on the object dictionary, we obtain the object names by linking the hand-interacted objects with their names.
First, we sample frames at 10 fps from an experiment video.
We process the sampled video segment using hand objects and Micro QR Code detectors.
Object names are obtained if the objects are positive (i.e., successfully detected by both modules).
The negative objects are fed into ResNet-50 to extract their feature vectors $\mathbf{n}$, and then we compute the cosine similarity between the obtained vector and mean vectors $\hat{\mathbf{p}}_{name}$ in the object dictionary.
The key with the highest similarity is assigned as the label of the negative objects.
Finally, the obtained object labels from the positive and negative objects are assigned to the video segments.

\begin{table}[t]
    \centering
    \caption{Recall of Micro QR Codes detection. Note that we do not use Micro QR Codes measuring 1cm$^2$ in the experiment ``making agarose gel'' because the experiment does not involve handling small equipment.}
    \label{tab:microQR_acc}
    \scalebox{0.9}{
    \begin{tabular}{l|rrrr}
    \hline
                       & \multicolumn{1}{c}{1cm} & \multicolumn{1}{c}{2cm} & \multicolumn{1}{c}{3cm} & \multicolumn{1}{c}{total} \\
    \hline
    DNA extraction     & 66.6                    & 50.0                    & 75.0                    & 73.6                      \\
    electrophoresis    & 55.5                    & 66.6                    & 50.0                    & 53.3                      \\
    making agarose gel & -                       & 66.6                    & 64.7                    & 65.2                      \\
    DNA purification      & 30.0                    & 55.5                    & 91.6                    & 61.2                      \\ \hline
    total              & 45.4                    & 58.3                    & 67.8                    & 60.9                  \\
    \hline
    \end{tabular}
    }
\end{table}

\begin{figure}[t]
    \centering
    \includegraphics[width=\linewidth]{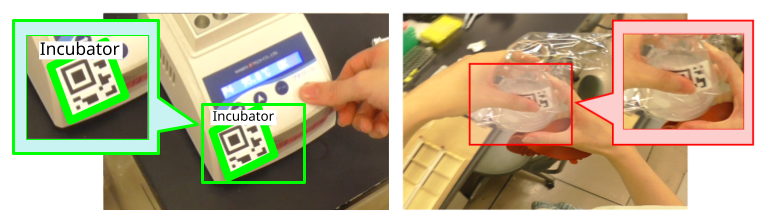}
    \caption{Success case (left) and failure case (right) of the Micro QR Code detection.}
    \label{fig:ex_microqr}
\end{figure}

\subsection{Evaluation}
\vspace{-0.5em}
\label{ssec:labeling_eval}
To assess the proposed object labeling method, we investigated the performance of the Micro QR Code detector in three steps.
First, we randomly sampled frames containing Micro QR Codes from experiment videos.
Then, these frames were processed through the hand object detector to extract only the frames where the objects were in contact with hands.
Based on this, we collected at least three frames per reagent or experimental equipment.
Finally, we applied the Micro QR Code detector to the collected frames and computed the size-wise detection performance for each experiment.
\Cref{tab:microQR_acc} indicates that Recall increases as the size increases in general, but the experiments of electrophoresis and making agarose gel do not show this tendency.
This is because the researcher manipulates the objects rapidly in these two experiments, frequently causing occlusion and motion blur (\Cref{fig:ex_microqr}).

\begin{table*}[t]
    \centering
    \caption{The step localization results for each biochemical experiment. StepFormer~\cite{stepformer} corresponds to the case of $\lambda=0$.}
    \label{table:result}
    \resizebox{0.95\textwidth}{!}{
        \begin{tabular}{|lccc|cccc|cccc|cccc|cccc|cccc|}
            \hline
            \multirow{2}{*}{Method} & \multirow{2}{*}{$\lambda$} & \multirow{2}{*}{QR} & Object & \multicolumn{4}{c|}{DNA extraction} & \multicolumn{4}{c|}{electrophoresis} & \multicolumn{4}{c|}{making agarose gel} & \multicolumn{4}{c|}{DNA purification} \\
            & & & labeling & MoF & Prec. & Rec. & tIoU & MoF & Prec. & Rec. & tIoU & MoF & Prec. & Rec. & tIoU & MoF & Prec. & Rec. & tIoU \\
            \hline
            \text{StepFormer}~\cite{stepformer} & & $\times$ & $\times$ & 56.2 & 61.7 & 60.9 & 46.5 & 41.9 & 45.6 & 47.1 & 33.4 & 30.5 & 28.3 & 28.8 & 21.3 & 24.7 & 27.1 & 28.0 & 15.8 \\
            \hline
            QR (ours) & 0.5 & $\checkmark$ & $\times$ & 50.5 & 53.8 & 57.5 & 40 & \textbf{55.4} & \textbf{60.1} & \textbf{59.4} & \textbf{43.4} & 27.9 & 27.1 & 29.1 & 16.2 & 27.2 & 24.4 & 29.1 & 16.1 \\
            \ +Labeling (ours) & 1.0 & $\checkmark$ & $\checkmark$ & 40.7 & 46.8 & 48.7 & 32.3 & 43.1 & 45.5 & 49.0 & 32.8 & 33.4 & \textbf{36.5} & \textbf{42.2} & \textbf{25.7} & 30.2 & 27.6 & 27.5 & 15.8 \\
            \ +Labeling (ours) & 0.5 & $\checkmark$ & $\checkmark$ & \textbf{58.9} & \textbf{63.8} & \textbf{67.0} & \textbf{50.5} & 49.4 & 53.3 & 53.2 & 37.9 & \textbf{34.9} & 31.5 & 35.2 & 20.8 & \textbf{30.7} & \textbf{31.1} & \textbf{32.6} & \textbf{19.0} \\
            \hline
        \end{tabular}
    }
\end{table*}

\section{Application: Step Localization}
\vspace{-0.5em}
\label{sec:steplocalization}
Based on BioVL-QR, we tackled the task of step localization to demonstrate the effectiveness of Micro QR Codes and our object labeling method.
In this section, we first formulate the problem of step localization (\Cref{ssec:problem}). Next, we introduce our approach to solving step localization (\Cref{ssec:approach}). Then, we present the detailed setup (\Cref{ssec:exp_setting}) and the results of our experiments (\Cref{ssec:Results}).

\subsection{Problem Formulation}
\vspace{-0.5em}
\label{ssec:problem}
We addressed the task of step localization, where each step in a procedural text is localized to a video segment~\cite{stepformer}. Given a video consisting of $N$ frames $(f_1, f_2, \ldots, f_N)$, and a textual sequence comprising $K$ steps $(s_1, s_2, \ldots, s_K)$, the objective is to predict segments that correspond to the steps. Each segment is represented as a pair of start and end frames $(f^{start}_i, f^{end}_i)  (1 \leq i \leq K)$. We assume that the steps do not overlap each other and are performed in the order in which they appear in the text.

\subsection{Our Approach}
\vspace{-0.5em}
\label{ssec:approach}
\textbf{Overview.}
Our approach is based on the existing step localization approach StepFormer~\cite{stepformer} and incorporates Micro QR Code information obtained by our labeling method into it.
StepFormer accepts video frame embeddings as inputs, so we add the embeddings of object names to them.
We use the pre-trained EgoVLPv2~\cite{egovlpv2} to embed egocentric videos and texts into a shared latent space. 

\noindent\textbf{StepFormer.}
For step localization upon the video and text features, we use StepFormer~\cite{stepformer}. StepFormer is a Transformer decoder model~\cite{Transformer} designed to be trained in a self-supervised manner~\cite{stepformer}. In our approach, we pre-train StepFormer on Ego4D and then fine-tune it on BioVL-QR.

\noindent\textbf{Micro QR Codes.}
Micro QR Code information is integrated into StepFormer using the following method. 
Using our object labeling method (see \Cref{sec:labeling}), we obtain the frame-wise names of $k$ objects, $(n_{i,1}, n_{i,2}, \dots, n_{i,k})$, that are touched in a given frame $i$. Each object name $n_{i,j}$ is encoded using the text encoder of EgoVLPv2 to produce embeddings $(\mathbf{e}_{i,1}, \mathbf{e}_{i,2}, \dots, \mathbf{e}_{i,k})$. The embeddings $(\mathbf{e}_{i,1}, \mathbf{e}_{i,2}, \dots, \mathbf{e}_{i,k})$ are summed as $\mathbf{E}_{i} = \sum_{j=1}^{k} \mathbf{e}_{i,j}$.
The aggregated embedding $\mathbf{E}_{i}$ is scaled by a constant $\lambda$ to adjust the influence of the object label names, resulting in $\lambda \mathbf{E}_{i}$. The scaled embedding $\lambda \mathbf{E}_{i}$ is then added to the video embedding $\mathbf{F}_{i}$ of the same frame to produce the enhanced frame embedding $\mathbf{V}_{i} = \mathbf{F}_{i} + \lambda \mathbf{E}_{i}$.
The enhanced video embeddings $\mathbf{V} = (\mathbf{V}_{1}, \mathbf{V}_{2}, \dots, \mathbf{V}_{s})$ are used as input instead of the original frame embeddings $\mathbf{F} = (\mathbf{F}_{1}, \mathbf{F}_{2}, \dots, \mathbf{F}_{s})$, where $s$ represents the number of frames in the video. Integrating object names into video embeddings is expected to mitigate the difficulty of localizing steps that involve visually indistinguishable objects.

\subsection{Experimental Settings}
\vspace{-0.5em}
\label{ssec:exp_setting}
\textbf{Baseline.}
We use StepFormer~\cite{stepformer} fine-tuned on BioVL-QR without incorporating Micro QR Code information. Except for the exclusion of this information, all other conditions remain the same as in the original paper.

\noindent\textbf{Implementation details.}
The step slots in StepFormer are fixed at $K = 32$. The number of epochs for fine-tuning is set to 60. The percentile drop cost of Drop-DTW~\cite{drop-dtw} is fixed at 0.75. We conducted experiments with $\lambda=0.5$ and $\lambda=1.0$.

\noindent\textbf{Data splits.}
We use the same set of videos for both training and testing.
Since StepFormer is trained in a self-supervised manner, it does not require step localization annotations, so there is no risk of data leakage.
Additionally, since the equipment, reagents, and containers vary depending on the biochemical experiment, we aim to develop a model specialized for the recorded videos rather than creating a generic one.

\noindent\textbf{Evaluation metrics.}
Following previous studies~\cite{stepformer, finebio}, we use MoF (Mean over Frames), precision, recall, and tIoU (temporal Intersection over Union). 
MoF is the ratio of frames correctly predicted as belonging to their ground-truth categories. 
Precision is the ratio of correctly predicted frames to all frames predicted for a specific step. 
Recall is the ratio of correctly predicted frames to all frames that actually belong to the step.
tIoU is the ratio of the intersection of frames belonging to a specific step and frames predicted as belonging to that step to their union.

\subsection{Results}
\vspace{-0.5em}
\label{ssec:Results}
\Cref{table:result} presents the evaluation results for each biochemical experiment in BioVL-QR. As we expected, the method using Micro QR Codes ($\lambda=0.5$) and object labeling method significantly outperforms the Vanilla StepFormer. Taking the average of the entire task, it improves MoF by $5.1\%$, Precision by $4.2\%$, Recall by $5.8\%$, and tIoU by $2.8\%$. These results suggest the usefulness of attaching Micro QR codes to objects and the object labeling method.
On the other hand, it can be observed that the performance does not improve significantly when $\lambda=1.0$ compared to when $\lambda=0.5$. This result suggests that overly relying on object-label information obtained from Micro QR Codes is sub-optimal. While the object label information is helpful for step localization, focusing excessively on the label information can potentially degrade task performance. Fusing both Micro QR Codes and visual information is essential for step localization.

\noindent\textbf{Ablation study.} We conducted step localization using only Micro QR Codes without the object labeling method.
As shown in \Cref{table:result}, compared to the method using only Micro QR Codes, the object labeling method ($\lambda=0.5$) led to performance improvements in three out of four biochemical experiments.
However, performance decreased only in electrophoresis.
This is likely because, in electrophoresis, the object labeling method correctly detects the micropipette. However, because the micropipette is not mentioned in the protocol, it does not serve as a cue for aligning visual and textual information and thus does not contribute to step localization. As a result, fusing the uninformative micropipette with visual representations leads to degraded performance.
Additionally, when comparing the method using only Micro QR Codes to the baseline, we observed that its performance was inferior.
This ablation study indicates that using only Micro QR Codes is insufficient, and combining them with the object labeling method improves performance in most cases.

\begin{figure}[t]
    \centering
    \includegraphics[width=\linewidth]{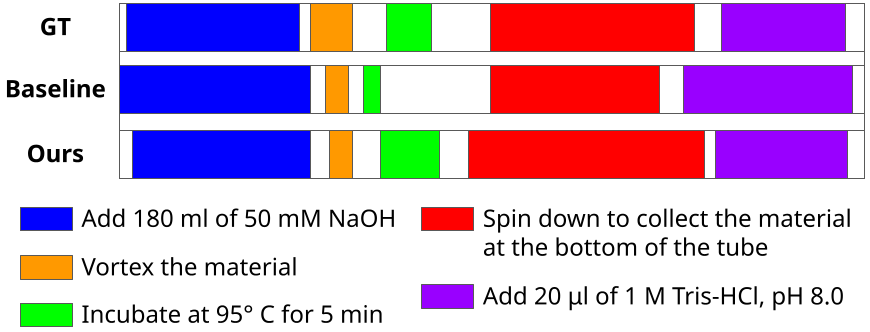}
    \caption{Sample of step localization for DNA extraction.}
    \label{fig:qualitative_result}
\end{figure}

\Cref{fig:qualitative_result} presents the qualitative results of step localization for DNA extraction.
In the baseline, the segment corresponding to the third step, ``Incubate at 95°C for 5 min'', could not be accurately localized.
However, in our approach, this step was correctly localized. This improvement is due to the use of micro QR Codes and the object labeling method, which enabled the identification of the incubator in the video (\Cref{fig:ex_microqr}).

\section{Conclusion}
\vspace{-0.5em}
\label{sec:conc}
We introduce BioVL-QR, a biochemical vision-and-language dataset containing videos, protocols, and alignment annotations.
Based on BioVL-QR, we conducted the task of step localization.
The experimental results show that our approach improves the understanding of biochemical videos.

\vspace{0.5em}
\begin{center}
{\large\bf Acknowledgements}
\end{center}
\vspace{-0.5em}
This work was supported by JSPS KAKENHI Grant Numbers 25K21274 and Joint Usage/Research Center for Interdisciplinary Large-scale Information Infrastructures (JHPCN) in Japan (Project ID: jh241007).

\vfill\pagebreak

\bibliographystyle{IEEEbib}
\bibliography{refs}

\begin{thebibliography}{10}

\bibitem{baker2016nature}
M.~Baker,
\newblock ``1,500 scientists lift the lid on reproducibility,''
\newblock {\em Nature}, vol. 533, pp. 452--454, 2016.

\bibitem{naim-2015}
I.~Naim, Y.~C. Song, Q.~Liu, L.~Huang, H.~Kautz, J.~Luo, and D.~Gildea,
\newblock ``Discriminative unsupervised alignment of natural language instructions with corresponding video segments,''
\newblock in {\em NAACL-HLT}, 2015, pp. 164--174.

\bibitem{finebio}
T.~Yagi, M.~Ohashi, Y.~Huang, R.~Furuta, S.~Adachi, T.~Mitsuyama, and Y.~Sato,
\newblock ``Finebio: A fine-grained video dataset of biological experiments with hierarchical annotation,''
\newblock {\em arXiv preprint arXiv:2402.00293}, 2024.

\bibitem{stepformer}
N.~Dvornik, I.~Hadji, R.~Zhang, K.~G. Derpanis, R.~P. Wildes, and A.~D. Jepson,
\newblock ``Stepformer: Self-supervised step discovery and localization in instructional videos,''
\newblock in {\em CVPR}, 2023, pp. 18952--18961.

\bibitem{biovl2}
T.~Nishimura, K.~Sakoda, A.~Ushiku, A.~Hashimoto, N.~Okuda, F.~Ono, H.~Kameko, and S.~Mori,
\newblock ``Biovl2: An egocentric biochemical video-and-language dataset,''
\newblock {\em Journal of Natural Language Processing}, vol. 29, pp. 1106--1137, 2022.

\bibitem{probio}
J.~Cui, Z.~Gong, B.~Jia, S.~Huang, Z.~Zheng, J.~Ma, and Y.~Zhu,
\newblock ``Probio: a protocol-guided multimodal dataset for molecular biology lab,''
\newblock {\em NeurIPS}, vol. 36, 2024.

\bibitem{microqr}
{ISO/IEC 18004:2024},
\newblock ``Information technology -- {{Automatic}} identification and data capture techniques -- {{QR}} code bar code symbology specification,''
\newblock Standard, International Organization for Standardization, 2024.

\bibitem{hobj}
D.~Shan, J.~Geng, M.~Shu, and D.~Fouhey,
\newblock ``Understanding human hands in contact at internet scale,''
\newblock in {\em CVPR}, 2020, pp. 9866--9875.

\bibitem{HoloAssist}
X.~Wang, T.~Kwon, M.~Rad, B.~Pan, I.~Chakraborty, S.~Andrist, D.~Bohus, A.~Feniello, B.~Tekin, F.~V. Frujeri, N.~Joshi, and M.~Pollefeys,
\newblock ``Holoassist: an egocentric human interaction dataset for interactive ai assistants in the real world,''
\newblock in {\em ICCV}, October 2023, pp. 20270--20281.

\bibitem{bhateja2024robotic}
C.~Bhateja, D.~Guo, D.~Ghosh, A.~Singh, M.~Tomar, Q.~Vuong, Y.~Chebotar, S.~Levine, and A.~Kumar,
\newblock ``Robotic offline rl from internet videos via value-function learning,''
\newblock in {\em ICRA}, 2024, pp. 16977--16984.

\bibitem{ego4d}
K.~Grauman, A.~Westbury, E.~Byrne, Z.~Chavis, A.~Furnari, R.~Girdhar, J.~Hamburger, H.~Jiang, M.~Liu, X.~Liu, M.~Martin, T.~Nagarajan, I.~Radosavovic, S.~K. Ramakrishnan, F.~Ryan, J.~Sharma, M.~Wray, M.~Xu, E.~Z. Xu, C.~Zhao, S.~Bansal, D.~Batra, V.~Cartillier, S.~Crane, T.~Do, M.~Doulaty, A.~Erapalli, C.~Feichtenhofer, A.~Fragomeni, Q.~Fu, A.~Gebreselasie, C.~Gonzalez, J.~Hillis, X.~Huang, Y.~Huang, W.~Jia, W.~Khoo, J.~Kolar, S.~Kottur, A.~Kumar, F.~Landini, C.~Li, Y.~Li, Z.~Li, K.~Mangalam, R.~Modhugu, J.~Munro, T.~Murrell, T.~Nishiyasu, W.~Price, P.~R. Puentes, M.~Ramazanova, L.~Sari, K.~Somasundaram, A.~Southerland, Y.~Sugano, R.~Tao, M.~Vo, Y.~Wang, X.~Wu, T.~Yagi, Z.~Zhao, Y.~Zhu, P.~Arbelaez, D.~Crandall, D.~Damen, G.~M. Farinella, C.~Fuegen, B.~Ghanem, V.~K. Ithapu, C.~V. Jawahar, H.~Joo, K.~Kitani, H.~Li, R.~Newcombe, A.~Oliva, H.~S. Park, J.~M. Rehg, Y.~Sato, J.~Shi, M.~Z. Shou, A.~Torralba, L.~Torresani, M.~Yan, and J.~Malik,
\newblock ``{{Ego4D}}: {{Around}} the {{World}} in 3,000 {{Hours}} of {{Egocentric Video}},''
\newblock in {\em CVPR}, 2022, pp. 18973--18990.

\bibitem{assembly101}
F.~Sener, D.~Chatterjee, D.~Shelepov, K.~He, D.~Singhania, R.~Wang, and A.~Yao,
\newblock ``Assembly101: A large-scale multi-view video dataset for understanding procedural activities,''
\newblock in {\em CVPR}, 2022, pp. 21096--21106.

\bibitem{youmakeup_2019}
W.~Wang, Y.~Wang, S.~Chen, and Q.~Jin,
\newblock ``{Y}ou{M}akeup: A large-scale domain-specific multimodal dataset for fine-grained semantic comprehension,''
\newblock in {\em EMNLP-IJCNLP}, 2019, pp. 5133--5143.

\bibitem{crosstask_2019}
D.~Zhukov, J.-B. Alayrac, R.~G. Cinbis, D.~Fouhey, I.~Laptev, and J.~Sivic,
\newblock ``Cross-task weakly supervised learning from instructional videos,''
\newblock in {\em CVPR}, 2019, pp. 3532--3540.

\bibitem{tvretrieval2020}
J.~Lei, L.~Yu, T.~L. Berg, and M.~Bansal,
\newblock ``Tvr: A large-scale dataset for video-subtitle moment retrieval,''
\newblock in {\em ECCV}, 2020, pp. 447--463.

\bibitem{gao2021fast}
J.~Gao and C.~Xu,
\newblock ``Fast video moment retrieval,''
\newblock in {\em ICCV}, 2021, pp. 1523--1532.

\bibitem{ResNet}
K.~He, X.~Zhang, S.~Ren, and J.~Sun,
\newblock ``Deep residual learning for image recognition,''
\newblock in {\em CVPR}, 2016, pp. 770--778.

\bibitem{egovlpv2}
S.~Pramanick, Y.~Song, S.~Nag, K.~Q. Lin, H.~Shah, M.~Z. Shou, R.~Chellappa, and P.~Zhang,
\newblock ``Egovlpv2: Egocentric video-language pre-training with fusion in the backbone,''
\newblock in {\em ICCV}, October 2023, pp. 5285--5297.

\bibitem{Transformer}
A.~Vaswani, N.~Shazeer, N.~Parmar, J.~Uszkoreit, L.~Jones, A.~N. Gomez, L.~u. Kaiser, and I.~Polosukhin,
\newblock ``Attention is all you need,''
\newblock in {\em NeurIPS}, 2017, pp. 5998--6008.

\bibitem{drop-dtw}
M.~Dvornik, I.~Hadji, K.~G. Derpanis, A.~Garg, and A.~Jepson,
\newblock ``Drop-dtw: Aligning common signal between sequences while dropping outliers,''
\newblock in {\em NeurIPS}, M.~Ranzato, A.~Beygelzimer, Y.~Dauphin, P.~Liang, and J.~W. Vaughan, Eds. 2021, vol.~34, pp. 13782--13793, Curran Associates, Inc.

\end{thebibliography}

\end{document}